\documentclass[sigconf,authorversion,nonacm]{acmart}

\usepackage{graphicx}
\usepackage{subfig}
\usepackage{latexsym}

\usepackage{amsfonts}
\usepackage{amsmath}
\usepackage{hyperref}

\usepackage{caption}

\DeclareMathAlphabet\mathbfcal{OMS}{cmsy}{b}{n}

\newcommand{\best}{\ensuremath{\mathsf{best}}}

\DeclareMathOperator*{\argmax}{argmax}

\newcommand{\halton}{\ensuremath{\mathsf{H}}}

\newcommand{\rf}{\ensuremath{\mathsf{RF}}}
\newcommand{\bb}{\ensuremath{\mathsf{BB}}}
\newcommand{\xb}{\ensuremath{\mathsf{XB}}}
\newcommand{\gp}{\ensuremath{\mathsf{GP}}}

\urldef{\blackit}\url{https://github.com/bancaditalia/black-it} 

\usepackage{tikz}
\usetikzlibrary{decorations.markings, arrows, arrows.meta,patterns.meta} 

\usepackage{algorithm}
\usepackage{algorithmic}
\usepackage{dsfont}
\usepackage{balance}

\begin{document}

\title{Reinforcement Learning for Combining Search Methods\\in the Calibration of Economic ABMs}


\author{Aldo Glielmo}
\authornote{Both authors contributed equally to this work. 
\\ This article was published in the \emph{Proceedings of the Fourth ACM International Conference on AI in Finance}, and is available also at \url{https://dl.acm.org/doi/abs/10.1145/3604237.3626889}.
\\ The views and opinions expressed in this paper are those of the authors and do not necessarily reflect the official policy or position of Banca d’Italia.
}
\email{aldo.glielmo@bancaditalia.it}
\orcid{0000-0002-4737-2878}

\author{Marco Favorito}
\authornotemark[1]
\email{marco.favorito@bancaditalia.it}
\affiliation{%
  \institution{Banca d'Italia}
  \country{Italy}
}
\orcid{0000-0001-9566-3576}

\author{Debmallya Chanda}
\email{debmallya.chanda@unicatt.it}
\affiliation{%
  \institution{Università Cattolica del Sacro Cuore and Universität Bielefeld}
  \country{Italy and Germany}
  }
\orcid{0000-0003-3067-739X}
% \affiliation{%
%   \institution{Universität Bielefeld}
%   \country{Germany}
%   }

\author{Domenico Delli Gatti}
\email{domenico.delligatti@unicatt.it}
\affiliation{%
  \institution{Università Cattolica del Sacro Cuore}
  \country{Italy}
}
\orcid{0000-0001-8819-090X}

\renewcommand{\shortauthors}{Glielmo, Favorito, Chanda, Delli Gatti}

\begin{abstract}
Calibrating agent-based models (ABMs) in economics and finance typically involves a derivative-free search in a very large parameter space. 
In this work, we benchmark a number of search methods in the calibration of a well-known macroeconomic ABM on real data, and further assess the performance of "mixed strategies" made by combining different methods. 
We find that methods based on random-forest surrogates are particularly efficient, and that combining search methods generally increases performance since the biases of any single method are mitigated.
Moving from these observations, we propose a reinforcement learning (RL) scheme to automatically select and combine search methods on-the-fly during a calibration run.
The RL agent keeps exploiting a specific method only as long as this keeps performing well, but explores new strategies when the specific method reaches a performance plateau. 
The resulting RL search scheme outperforms any other method or method combination tested, and does not rely on any prior information or trial and error procedure.
\end{abstract}







\maketitle

\section{Introduction and literature review}

The last decades have witnessed a consistent growth of the reach and scope of agent-based models (ABMs) in economics and finance, certainly also as a consequence of continuing improvements in the computer hardware and software that form the foundation over which ABMs are designed and used \cite{axtell2022agent}.
ABMs have also become mature enough that they have seen adoption and usage within central banks and other financial institutions for specific tasks \cite{turrell2016agent,plassard2020making}. 
A particularly noteworthy application domain is the modelling of the housing market, pioneered by Bank of England \cite{baptista2016macroprudential} and later studied by many other central banks \cite{cokayne2019effects,catapano2021macroprudential,carro2022could,mero2022ahigh}, and the macroeconomic model proposed in \cite{poledna2023economic} and recently adopted by Bank of Canada~\cite{hommes2022canvas}.
Other successful applications can be found in the modelling of financial stability~\cite{bookstaber2018agent,covi2019origin}, or of the banking sector~\cite{chan2017abba}.
Finally, ABMs for economic and financial markets are currently being investigated, extended, and possibly used, within JPMorgan Chase~\cite{ardon2021towards,vadori2022towards,ardon2023phantom}.

In spite of these success stories, ABMs still occupy a minor for modelling in economics and finance.
One fundamental reason behind ABMs' limited adoption is the overwhelming flexibility of such a modelling tool which, if handled incorrectly, can lead to widely different models of the same phenomenon and consequently to a narrow predictive power.

Rigorous calibration of ABMs via large amounts of real data is a promising path to address the problem of ABM flexibility by appropriately restricting it in data-driven and systematic manner~\cite{axtell2022agent}.
In fact, ABM calibration has a long history~\cite{fagiolo2007critical}, but interest in ABM calibration  has grown particularly in recent times of ever-increasing data abundance.
% 
% accurate model calibration is a long standing problem in the field of economic ABM development \cite{fagiolo2007critical}, that has grown particularly in recent times of ever-increasing data abundance.
% it is not a coincidence that interest in ABM calibration has closely paralleled the interest in ABM development, and has grown particularly in recent times of ever-increasing data abundance.
%
Historically, the problem has been approached mostly via the `method of simulated moments' \cite{gilli2003global,franke2009applying,grazzini2015estimation}, which involves minimising a measure of distance between summary statistics of real and simulated time series, while more recently, other approaches based on maximum likelihood or Bayesian statistics have been proposed and successfully tested \cite{grazzini2017bayesian, platt2021bayesian,dyer2022black,monti2023learning}.

A common challenge of all calibration frameworks is the need of efficiently searching for optimal parameter combinations in high-dimensional spaces, a problem made particularly arduous by the high computational cost of state-of-the-art ABM simulations.
This is why the use of several heuristic search methods has been proposed in the ABM literature.
Specifically, in~\cite{lamperti2018agent}, building on the work of~\cite{conti2010bayesian}, the authors propose the use of machine surrogates, specifically in the form of XG-boost regressors, to suggest promising parameter combinations by interpolating the results of previously computed ABM simulations. 
In \cite{angione2022using}, the authors expand on this idea and test the ability of several machine learning surrogate algorithms such as Gaussian processes, random forests and support vector machines, to reproduce ABM simulation data.
In \cite{platt2020comparison} the author instead proposes the use of particle swarm samplers \cite{kaveh2017particle,stonedahl2011genetic}, as well as the search heuristic of \cite{knysh2016blackbox}.

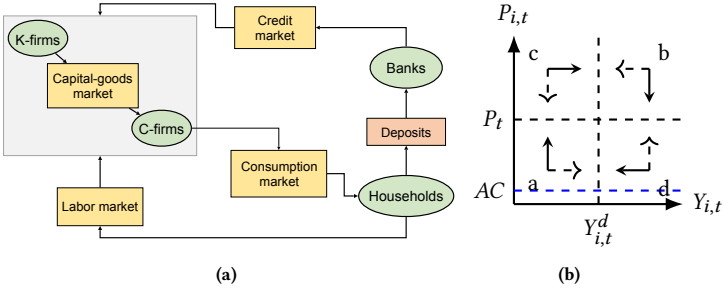
\begin{figure}[t!]
\advance\leftskip-1.2cm
    \subfloat[\centering]{{\includegraphics[trim={3.0cm 1.5cm 3.8cm 2.5cm},clip,width=6cm]{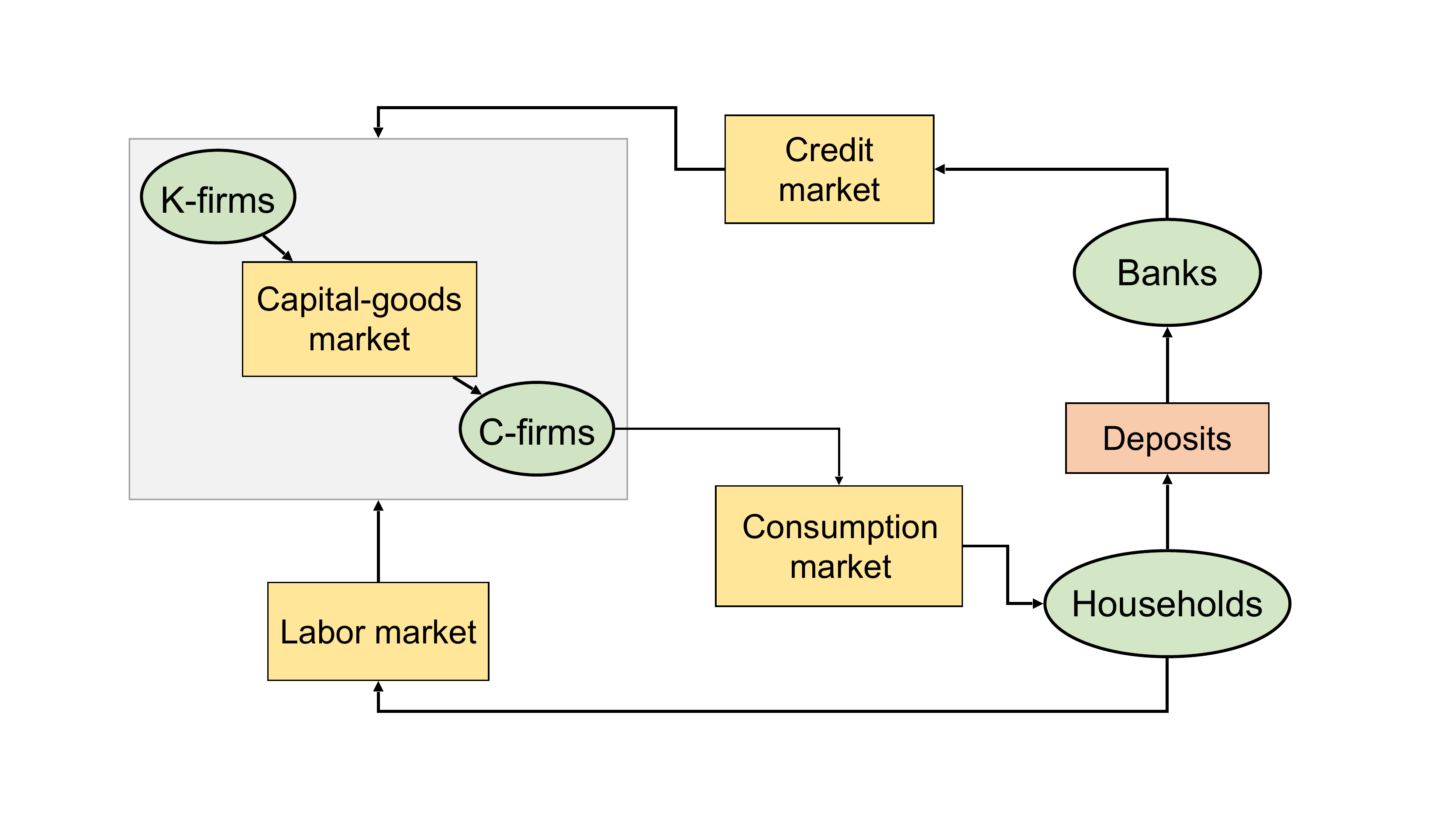}}}%
    \hspace{0cm}
    \subfloat[\centering]{{
    	\begin{tikzpicture}[thick, scale=0.45]
		\draw[-{Latex}] (0,0) -- (0,5) node[above] {$P_{i,t}$};
		\draw[-{Latex}] (0,0) -- (5,0) node[right] {$Y_{i,t}$};
		
		\draw [dashed] (2.5,0) node[below]{$Y_{i,t}^d$} -- (2.5,5) ;
		\draw [dashed] (0,2.5) node[left]{$P_t$} -- (5,2.5);
		
		\draw [draw=blue, dashed] (0,0.4) node[left]{$AC$}-- (5,0.4);
	
		\draw (1,4) node[above left]{c};   
		\draw (4,4) node[above right]{b};  
		\draw (1,1) node[below left]{a};  
		\draw (4,1) node[below right]{d};

		\draw [-stealth ](1,4) -- (2,4);
		\draw [dashed,-> ](1,4) -- (1,3);
		
		\draw [-stealth ](4,4) -- (4,3);
		\draw [dashed,-> ](4,4) -- (3,4);
		
		\draw [-stealth ](1,1) -- (1,2);
		\draw [dashed,-> ](1,1) -- (2,1);
		
		\draw [-stealth ](4,1) -- (3,1);
		\draw [dashed,-> ](4,1) -- (4,2);
	
	\end{tikzpicture}
 \hspace{-0.8cm}
 }}
    \caption{The CATS model. (a) An illustration of the agent classes of the model and their interactions. 
    Agent classes are represented in green ovals, interaction types are specified in rectangles, and markets are specified in yellow rectangles. 
    The directions of the arrows indicate the flow of the specific good e.g., consumption-goods are acquired by households from C-firms, while labour is acquired by firms from households. 
    (b) An illustration of the firms' decisions on the price-quantity space.
    Prices $P_{i,t}$ and quantities $Y_{i,t}$ of goods are updated following the 4 solid black arrows,
    % (representing Equations (4%\ref{eq:CATS-price-setting}
    % ) and (3%\ref{eq:CATS-quantity-setting}
    % ) of Appendix A)
    % , 
    and not the  dashed black arrows.
    The dashed blue line is the minimum price they can charge, corresponding to the average cost ($AC$) for production.
    }
    \label{fig:cats-model}
    \vspace{-0.2cm}
\end{figure}

In this work, we take a different view of the problem and test the performance of existing search strategies, on a common calibration task, and propose simple methods to combine them in mixed strategies to drastically boost calibration performance.
%
%
% \mar{- highlight the related works that combine multiple search methods in the same calibration process. E.g. the work \cite{knysh2016blackbox} is precisely this, although with a fixed sampler scheduling policy. Other works? Perhaps \cite{sazanovich_solving_nodate}?}\\
% \mar{We could say something along these lines: There are other works that explore the idea of combining different search methods in the same calibration process to obtain a more effective calibration...}
%
We test our methods one of the most well-known and studied macroeconomic ABMs~\cite{gatti2011macroeconomics,assenza2015,dawid2018agent}, often referred to as the \emph{CATS} (``Complex Adaptive Trivial System'') model.
Our contributions are as follows:
\vspace{-0.1cm}
\begin{itemize}
  \setlength\itemsep{-0.03em}
    \item We verify that the macroeconomic ABM considered can be efficiently calibrated to reproduce a variety of real time series.
    \item We find that methods based on random forest machine learning surrogates are particularly effective searchers in the highly non-convex and discretely-changing loss function induced by ABMs.
    \item We find that combining together different search methods almost always provides better overall performance, and propose this as a convenient heuristic to apply in the ABM calibration practice.
    \item We introduce a simple reinforcement-learning technique to automatically aggregate any number of search methods in a single mixed strategy, and demonstrate the superior performance of this approach with respect to naive aggregation strategies.
\end{itemize}
\vspace{-0.1cm}

In Section~\ref{sec:model-description} we overview the main features of the CATS model, 
% (an extended description is provided in Appendix
% A%~\ref{sec:long-model-description}
% ), 
in Section~\ref{sec:calibration-description} we describe the calibration technique considered and the search methods that we employ individually and in combination, in Section~\ref{sec:benchmarks} we describe our benchmarking experiments and the results obtained, in Section~\ref{sec:reinforcement-learning} we describe the reinforcement learning scheme we proposed to automatically combine existing methods, and demonstrate its performance, in Section~\ref{sec:validation} we verify that the calibrated model reproduces the target real data and that our findings hold well against changes in the model and in the loss function, in Section~\ref{sec:conclusion} we conclude.

%
%In the interest of reproducibility, the code and the data used to generate the key results of this work are available to download as supplementary material of the paper.

\section{Model illustration}
\label{sec:model-description}

The CATS model consists of four classes of interacting agents: households, final-goods producing firms (C-firms), capital producing firms (K-firms) and banks. 
Figure~\ref{fig:cats-model}a illustrates these classes of agents and the markets through which they interact, while Figure~\ref{fig:cats-model}b illustrates a distinctive feature of the model, i.e., the decision making operated by firms on price and quantity of goods to produce.
In the interest of space, and since this work focuses on the calibration of the model, we do not describe the details of the model here and refer the reader to \cite{assenza2015} for an in-depth exposition.

% but report them in Appendix 
% A%\ref{sec:long-model-description}
% . 
% We also refer the reader to \cite{assenza2015} for an in-depth exposition. 

\section{Calibration description}
\label{sec:calibration-description}

The calibration method we consider is composed of three main steps.
First, a search method (from now on also called a \emph{sampler}) suggests a set of parameters to explore, then a number of simulations are performed for each selected parameter, and finally a loss function is evaluated to measure the goodness of fit of the simulations with respect to the real time series.
Iterating these three steps allows finding parameters corresponding to progressively lower loss values, and the parameter corresponding to the lowest loss value can be considered optimal.

We follow the \emph{method of moments} paradigm~\cite{franke2009applying,chen2018estimation} and use the following loss function (often called \emph{distance} in the ABM literature) for all calibrations.
This takes the form
\begin{equation}
\label{eq:loss-function}
    L = \frac{1}{D} \sum_{d=1}^D \mathbf{g}_d^{\rm{T}} \mathbf{W}_d \mathbf{g}_d ,
\end{equation}
where $\mathbf{g}_d$ is the vector of difference between the real and the simulated moments of the one-dimensional time series $d$, and $D$ is the total number of dimensions in the multi-dimensional time series considered.
Different choices for the weighting matrices $\mathbf{W}_d$ have been proposed in the literature \cite{franke2009applying,franke2012structural}.
In this work we take the $\mathbf{W}_d$ matrices to be diagonal matrices with elements $(\mathbf{W}_d)_{ii}$ inversely proportional to the square of the real $i$-th moment of the one-dimensional time series $d$.
This choice guarantees that the same weight is given to all moments considered, independently of the different scales or units of measure that the different moments might have.
In essence, the loss function written in this way provides an estimate of the relative squared error between real and simulated moments.

Since we use a common loss function for all calibrations, the only difference between the calibration runs considered here is the choice of search method.
We consider the following five search methods, all of which are implemented in \emph{Black-it}~\cite{black_it}, an open source library for ABM calibration: 
%\blackit
~\footnote{\blackit}
%\footnote{\url{https://github.com/bancaditalia/black-it}}. 
%

\noindent
\textbf{Halton sampler $\ensuremath{\mathsf{(}} \halton \ensuremath{\mathsf{)}}$.}
This sampler suggests points according to the Halton series~\cite{halton1964algorithm,kocis1997computational}.
The Halton series is a low-discrepancy series providing a quasi-random sampling that guarantees an evenly distributed coverage of the parameter space.
As the method is very similar to a purely random search, we use it as a baseline for the more advanced search strategies analysed.

% The other three samplers considered are ``adaptive'' search methods meaning that they make use of the information coming from previously explored parameters in order to continue searching in the vicinity of the locally optimal parameters.

\noindent
\textbf{Random forest sampler $\ensuremath{\mathsf{(}} \rf \ensuremath{\mathsf{)}}$.}
This sampler is a type of machine learning surrogate sampler. It interpolates the previously computed loss values using a random forest classifier \cite{bajer2015benchmarking}, and it then proposes parameters in the vicinity of the lowest values of the interpolated loss surface.
%
%We use a random forest classifier with 500 independent estimators (``trees'') and use 10 classes chosen as the 10 quantiles of the distribution of evaluated losses.

\noindent
\textbf{XG-boost sampler $\ensuremath{\mathsf{(}} \xb \ensuremath{\mathsf{)}}$.}
This sampler is a machine learning surrogate sampler that interpolates loss values using an XG-Boost regression \cite{chen2016xgboost}, as proposed in \cite{lamperti2018agent}.
%
%We use a learning rate of 0.1, a   maximum tree depth of 5,        and 10 estimators.

\noindent
\textbf{Gaussian process sampler $\ensuremath{\mathsf{(}} \gp \ensuremath{\mathsf{)}}$.}
This sampler is a machine learning surrogate sampler that interpolates loss values using a Gaussian process regression \cite{conti2010bayesian,rasmussen2004gaussian}.
%
%We use a Matérn covariance function with $\nu = 5/2$ and with the lengthscale optimised at every iteration via maximum marginal likelihood.

\noindent
\textbf{Best batch sampler $\ensuremath{\mathsf{(}} \bb \ensuremath{\mathsf{)}}$.}
This sampler is a very essential type of genetic algorithm \cite{stonedahl2011genetic} that takes the parameters corresponding to the current lowest loss values and perturbs them slightly in a purely random fashion to suggest new parameter values to explore. 
%
%The random perturbation is specifically obtained by first selecting a random subset of dimensions, and then changing the parameter value along those dimensions uniformly but within a short range (plus/minus 0.006 in our case).
%

\begin{figure*}[!ht]
\centering
\advance\leftskip-1cm \includegraphics[width=.75\linewidth,trim={0.35cm 0 0.35cm 0},clip]{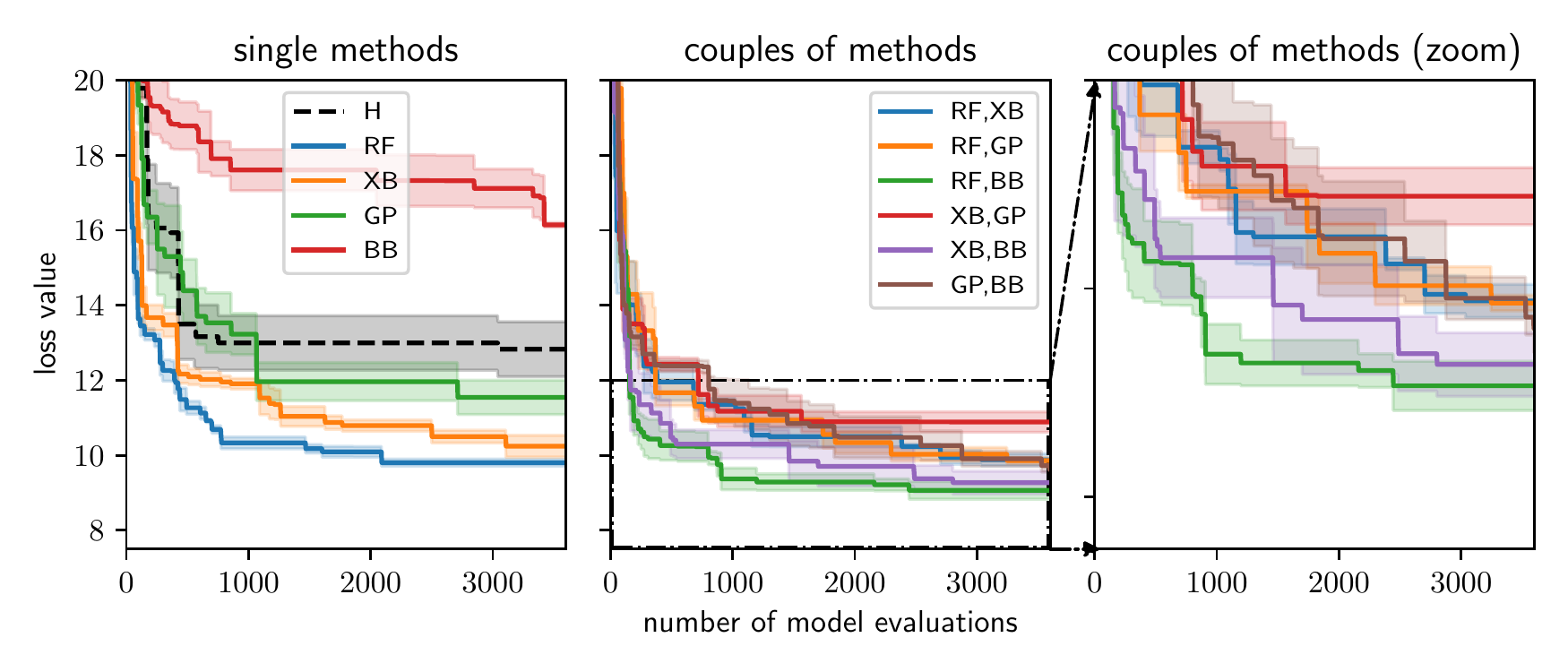}

\centering
\small
%\resizebox{0.98 \linewidth}{!}{
\begin{tabular}[b]{c | ccccc | cccccc }\hline
Method & H & RF & XB & GP & BB & RF,XB & RF,GP & RF,BB & XB,GP & XB,BB & GP,BB \tabularnewline
\hline 
\hline
Mean & 12.83  & \textbf{9.803} & 10.24 & 11.96 & 16.87 & 9.88 & 9.861 & \textbf{9.07} & 10.89 & 9.27 & 9.83 \tabularnewline
Std. Err. & 0.73 & 0.094 & 0.29 & 0.51 & 0.59 & 0.16 & 0.075 & 0.24 & 0.27 & 0.31 & 0.26 \tabularnewline
\hline
\end{tabular}
%}
\caption{
Top graphs: Loss as a function of the number of model evaluations for the single methods (left), and for couples of methods (middle and right).
Bottom table: Means and standard errors of the lowest losses achieved by the different strategies.
Note that these results are directly comparable with those shown in Figure~\ref{fig:convergence-rl} and discussed in the next section, as both $x$ and $y$ axes have identical ranges.
}
\label{fig:convergence}
\vspace{-0.2cm}
\end{figure*}

\section{Benchmarking experiments}
\label{sec:benchmarks}

\begin{table}[t]
\centering
%\footnotesize
\small
\begin{tabular}{llll}
\hline 
Param. & Description & Range  \tabularnewline
\hline 
\hline 
$\xi$ & Memory parameter in consumption & 0.5-1  \tabularnewline
$\chi$ & Wealth parameter in consumption & 0-0.5  \tabularnewline
$\rho$ & Quantity adjustment & 0-1  \tabularnewline
$\bar{\eta}$ & Price adjustment & 0-1  \tabularnewline
$\mu$ & Bank\textquoteright s gross mark-up & 1-1.5  \tabularnewline
$\zeta$ & Bank\textquoteright s leverage & 0-0.01  \tabularnewline
$\delta^k$ & Inventories depreciation rate & 0-0.5  \tabularnewline
$\gamma$ & Fraction of investing C-firms & 0-0.5  \tabularnewline
$\theta$ & Rate of debt reimbursement & 0-0.1  \tabularnewline
$\nu$ & Memory parameter in investment & 0-1  \tabularnewline
$t_w$ & Tax rate & 0-0.4  \tabularnewline
\hline
\end{tabular}
\caption{Parameter descriptions and their ranges.}
\label{tab:parameter-ranges}
\vspace{-0.8cm}
\end{table}

\noindent
\textit{Experiments preparation.}
Similarly to \cite{gatti2020rising}, we calibrate the model using the following 5 historical time series, representing the US economy from 1948 to 2019, downloaded from the FRED database~\cite{mccracken2016fred}: total output, personal consumption, gross private investment (all in real terms), the implicit price deflator and the civilian unemployment.
To make simulated and observed data comparable, we remove the trend component from the total output, consumption and investment using an HP filter \cite{ravn2002adjusting}; and we use simulated and observed price deflator to compute de-meaned inflation rates.
% In Appendix 
% B %\ref{appendix:transient-effects}
% we provide an example of the resulting time series, and verify that 300 simulation epochs are sufficient to equilibrate the model.
% % 
In Table \ref{tab:parameter-ranges} we list the 11 parameters considered for calibration and the specified ranges of variation.

\noindent
\textit{Experiments performed.}
Using the four samplers described in the previous section, we build 11 search methods as the 5 samplers taken individually, as well as the 6 combinations of any two non-baseline samplers.
For each search method, we perform 3 independent calibration runs.
Each calibration run consists of 3600 model evaluations, and for each parameter 5 independent simulations are performed to reduce the statistical variance of the loss estimate.
Each simulated series consists of 800 time-steps generated by running the model for 1100 time steps and discarding the first 300.
This makes up a total of almost 600000 simulations and more than 50 days of CPU time, which we were able to compress in less than two days by leveraging parallel computing both within and between calibrations.

\noindent
\textit{Results and discussion.}
Figure \ref{fig:convergence} reports the cumulative minimum loss achieved by the different sampling strategies as a function of the number of model evaluations performed.
The lines and the shaded areas indicate averages and standard errors over the 3 realisations of the experiment.
Single samplers are reported in the left graph, while couples of samplers are reported in the middle graph as well as --zoomed-- in the right graph.
The table at the bottom of the graphs reports the minimum loss achieved by the different methods.

\vspace{0.2cm}
\noindent
\textbf{Single methods.}
When samplers are taken in isolation, the random forest sampler ($\rf$) clearly outperforms all other methods, the XG-boost sampler ($\xb$) is the second best performing and the Gaussian process sampler ($\gp$) is substantially worse than the other two machine learning surrogate samplers.
The low performance of the $\gp$ sampler can be ascribed to the smoothness and regularity assumptions inherent in Gaussian process regression models, assumptions that are not present in random-forest or XG-boost models, and not suited to describe the roughed and complex loss landscape of ABM calibrations.
The best batch sampler ($\bb$) performs very poorly in isolation, and underperforms even in comparison with the baseline $\halton$ sampler.
This is not entirely surprising, since the $\bb$ sampler can only propose small perturbations around current loss minima and can thus easily remain stuck in one of the many local minima of the highly non-convex landscape typical of ABMs loss functions.

\vspace{0.2cm}
\noindent
\textbf{Couples of  methods.}
All methods, not just the poorly performing $\bb$ sampler, possess intrinsic sampling biases that in the long run can hinder their performances and make them converge to sub-optimal solutions.
We find that combining different methods in mixed strategies can strongly mitigate such biases and improve overall performance.
The effect can be observed in the second and third panel of Figure~\ref{fig:convergence}, by noticing that couples of methods, with the only exception of the `$\xb,\gp$' combination, always perform on par or better than the best single samplers ($\rf$ and $\xb$).
Interestingly, the best overall performances are achieved by coupling one machine learning surrogate sampler with the genetic $\bb$ sampler.
In light of the above discussion, we note that machine learning surrogate samplers and the $\bb$ sampler work in very different ways, and hence their combination can strongly diminish the respective sampling biases, while since machine learning surrogates all work in similar ways, their combination does not yield to comparable improvements. 
The $\rf,\bb$ and the $\xb,\bb$ combinations are particularly effective and achieve the lowest loss values.

% \paragraph{Triplets of methods.}
% %
% Among the triplets of methods considered, the combination $\halton,\rf,\bb$ deserves a special mention.
% %
% It achieves roughly the same loss values as the $\rf,\bb$ combination, but it converges to it significantly faster (using 1000 model calls rather than 2000, note the log-scale on the $x$-axis).
% %
% 
% \vspace{0.5cm}

To summarise, our results show that the $\rf$ and $\xb$ samplers are particularly well suited to efficiently search in the parameter space of ABMs. 
%
% This is in agreement with the claims in~\cite{lamperti2018agent}, which more generally concern the efficiency of machine learning surrogate modelling of ABM loss functions.
%
The success of the $\rf$ and $\xb$ samplers can be ascribed to the
ability to correctly approximate high dimensional and possibly discontinuous functions with no regularities.
However, the performance of the $\rf$ and $\xb$ samplers can be significantly improved if they are used in combination with the $\bb$ sampler.
%

% The results presented so far can already offer useful guidance for researchers interested in calibrating medium and large scales ABMs, as they provide an easy recipe to boost calibration efficiency by simple alternation of existing search methods.
% %
In the next section, we move a step forward and consider the combination of multiple methods in more general terms, without limiting ourselves to the simplest scenario of a ``round-robin'' selection.

% samplers such as in the $\rf,\halton$ or in the $\rf,\bb$ and in particular in $\halton,\rf,\bb$ sampler, in agreement with what found in \cite{black_it} and justifying the choice of operated in \cite{catapano2021macroprudential} for the calibration of a large scale ABM of the real estate sector.
%
% In fact, the $\rf$ sampler is based on a machine learning approximation of the real loss landscape and, to the extent to which the approximation can be locally inexact, the sampler can remain stuck in sampling sub-optimal minima.
% %
% This is why coupling it with other methods can improve the overall balance between \emph{exploration} and \emph{exploitation} and give rise to empirically better calibration strategies.
%

\section{Reinforcement learning experiments}
\label{sec:reinforcement-learning}
The results of the benchmarks presented in Section~\ref{sec:benchmarks} show that the combination of different types of sampling methods can be beneficial for the calibration process even when we naively alternate the available sampling methods during the course of a calibration. 
This suggests that the investigation of different --and more flexible-- scheduling policies of search methods could bring to even more efficient calibrations.

In particular, it is desirable that the chosen scheduling policy shows some form of \emph{adaptivity}, i.e. that is able to choose  the sampling method with more chances to sample a good parameter vector, taking into account the progress of the calibration process. 
To achieve this goal, we frame the ABM calibration problem as a reinforcement learning (RL) problem where the decision-maker (the \emph{agent}) has to find a good policy such that it chooses the most promising search method, where ``promising'' is related to the chances of sampling a parameter that improves the value of the loss. 
The decision-maker receives feedback for its choice in the form of a reward signal computed from the sampled loss function values. 
This is what makes the scheduling policy adaptive: search methods that more often provide loss improvements are more rewarding from the decision-maker perspective, and they have more chances of being chosen in the next calibration step;
on the other hand, whenever a search method does not show to be rewarding anymore,
%e.g. because it is too explorative with respect to the current phase of the calibration process, 
then the decision-maker can detect this and switch the preference to another search method.
%

% Borrowing terminology from control theory~\cite{richard2008modern}, fixed scheduling policies, such as the naive samplers' combinations explored in the previous section, are \emph{open-loop}, i.e. they do not change regardless of how a search method is performing; instead, RL-based scheduling policies are \emph{closed-loop}, because they receive and process the feedback coming from the calibration process, possibly reacting to such feedback by changing the preferred sampling method.

Specifically, we frame the calibration process as a \emph{multi-armed bandit (MAB)} problem \cite{katehakis1987multi,weber1992gittins,auer2002finite,berry1985bandit,gittins2011multi,lattimore2020bandit}.
% 
% Informally, in a MAB problem, a fixed limited set of resources must be allocated between competing (alternative) choices in a way that maximizes their expected gain, where the actual outcome of each choice is unknown a priori.
%
This is a classic reinforcement learning problem that exemplifies the \emph{exploration–exploitation trade-off dilemma}~\cite{sutton2018reinforcement}.
The challenge for the agent is to simultaneously attempt to acquire new knowledge by ``exploring'' different \emph{actions} and optimise their decisions based on existing knowledge by ``exploiting'' actions that have been estimated to be rewarding.  
% \todo{cite MAB applications}
%
% The name comes from imagining a gambler at a row of slot machines (sometimes known as ``one-armed bandits"), who has to decide which machines to play, how many times to play each machine and in which order to play them, and whether to continue with the current machine or try a different machine. ALDO: removed as it seems too much elmentary
%Continuing on this analogy,
%
We define the different sampling methods as the actions available for the agent, and loss improvements as the reward signals.
More formally, we define the reward at time $t$ as the fractional improvement achieved over the previous best loss
\begin{equation}
\label{eq:mab-reward}
R_t = \max \{ 0, \frac{ L_{\best,t-1} - L_t }{L_{\best,t-1}} \}
\end{equation}
where $L_t$ is the loss obtained for the simulations sampled at time $t$, and $L_{\best,t-1}$ is the best loss sampled up to time $t-1$. 
Note that $R_t$ is a random variable, because $L_t$ depends on the simulated time series outputted by the ABM, and the chosen parameter vector.
As in most of the MAB problems, the goal for the agent is to maximize the cumulative sum of rewards
\begin{equation}
\label{eq:mab-objective}
    S_N = \sum\limits_{t=1}^N R_t,
\end{equation}
where $N$ is the number of calibration steps. 
%
% It is easy to see that the maximization of the cumulative sum of rewards (\Cref{eq:mab-objective}) is equivalent to the minimization of the calibration loss function (\Cref{eq:loss-function}). ALDO: removed this as it's probably not true if we take percentage improvements in the reward definition

Differently from the usual MAB setting, the reward probability distributions associated to each available sampler are obviously \emph{non-stationary}~\cite{AuerCFS02}, and in fact they change drastically during the course of the calibration. 
As an example, consider that at end of a calibration all methods --even the best ones-- stop providing any improvement in the loss, and hence the reward distributions become progressively more peaked around zero.
%
% Non-stationarity is the most general assumption one can make over the behaviour of reward probability distributions in MABs~\cite{AuerCFS02} and, in our case, the non-stationarity assumption is required from of the lack of knowledge on both the ABM and the samplers' behaviours. 
% %
% As an example where non-stationarity of rewards can happen, consider a calibration setup with a random uniform sampler $\sampler$ over the parameter space, with best known loss $\ell_\best$, and the expected loss improvement gained by using the random sampler at time $t$, $\mu_t$. 
% %
% Suppose that in the calibration step $t$, a better parameter vector is sampled, yielding a new best loss, $\ell_\best' < \ell_\best$. 
% %
% Which is the expected reward of the random sampler at the next time step, $\mu_{t+1}$? Since the sampler is uniform, and since the set of parameters that yields a loss $\ell$ such that $\ell_\best' < \ell \le \ell_\best$ does not contribute anymore to the reward expectation after time step $t$, we necessarily have $\mu_{t+1} \le \mu_t$. 
% %
% Therefore, the expected reward of the sampler has changed across time steps, which is a non-stationary behaviour of the reward distribution.

% The reader might wonder why other RL approaches have not been considered, e.g. the classical approaches that model the environment as a Markov Decision Processes~\cite{sutton2018reinforcement}, where the decisions of the agent can be taken starting from a set of features, thus allowing for more informed decisions.
% %

The MAB is a very simple framework for RL problems, that are more generally modelled as Markov Decision Processes (MDPs)~\cite{sutton2018reinforcement}.
However, their simplicity is precisely what makes MAB better suited for our context than other approaches.
Indeed, as MAB algorithms focus on finding the best action at each step rather than learning the entire environment, they are much more sample efficient.
In the ABM calibration context, simulations are typically very expensive, and consequently the sample efficiency of the learning method is of paramount importance.
%
% Additionally, MAB algorithms can adjust their exploration-exploitation trade-off on-the-fly, which allows them to converge to the optimal action more quickly than MDPs, which often require a fixed exploration schedule. \ald{[do we have a citation for this? Otherwise we should remove this sentence]}
% %
% This is also important as we require the policy to adapt on-the-fly during a calibration.

% %
% In MDPs, an agent must explore the entire state space in order to learn the optimal policy, whereas in MAB algorithms, the agent only needs to explore the different actions in order to learn the optimal one.
% %
% Additionally, MAB algorithms can adjust their exploration-exploitation trade-off on-the-fly, which allows them to converge to the optimal action more quickly than MDPs, which often require a fixed exploration schedule.
% %
% Since in the ABM calibration context we usually consider model simulations as being very expensive, the sample efficiency of the learning method is of paramount importance.

% However, MDPs are more powerful than multi-armed bandit algorithms because they can handle environments with multiple states and more complex dynamics.

%In the following, we consider different variants of the MAB problem in order to train good scheduling policies for our calibration problem. 
%
%We consider two learning paradigms: \emph{offline-learning}, where the agent learns from previously executed calibrations with classical sampling methods, and \emph{online-learning}, where the agent actively influences the calibration process, and learns from the received feedback.
%
In the following, we test our MAB framework in two experiments.
First, in the \emph{offline-learning} experiments, we let the agent learn from the previously executed calibrations of Section~\ref{sec:benchmarks}.
%and extract useful insights from the learned best policies.
%
Then, in the \emph{online-learning} experiments, we let the agent interact with the environment and optimise its policy on-the-fly during each calibration.

\begin{table}[t]
\centering
%\footnotesize
\small
\begin{tabular}{c | cccccc}
\hline 
Sampler \textbackslash\ Context  & sing. samp. & glob. & high $L_{\best,t}$ & low $L_{\best,t}$ \tabularnewline
\hline 
\hline 
$\rf$     & \textbf{0.25} & \textbf{0.27} & \textbf{1.3}  & 0.052  \tabularnewline
$\xb$     & 0.23 & 0.23 & 0.61 & 0.033 \tabularnewline
$\gp$     & 0.21 & 0.17 & 0.26 & 0.068  \tabularnewline
$\bb$     & 0.11 & 0.23 & 0.28 & \textbf{0.18}   \tabularnewline
$\halton$ & 0.20 & 0.20 & 0.24 & N.A.  \tabularnewline
\hline 
\end{tabular}
\caption{The estimated $Q$ functions for the different search methods and under different contexts. (sing. samp.) uses only single sample calibrations, (glob.) uses all calibrations, (high $L_{\best,t}$) uses all calibration but only actions taken when the loss is \emph{above} the median loss, and (low $L_{\best,t}$) uses all calibration but only actions taken when the loss is \emph{below} the median loss. Results are reported on a scale of $10^{-3}$.} 
\label{tab:rl-experiment-offline}
\vspace{-0.8cm}
\end{table}

%\begin{figure*}[t]
%\centering
%\advance\leftskip-1cm \includegraphics[width=.64\linewidth,trim={0.35cm 0 0.35cm 0},clip]{figures/convergence_1.pdf}
%\hfill
%\centering
%\includegraphics[width=0.30
%\linewidth,trim={0 0 0 0},clip]{figures/actions.png}
%\small
%\begin{tabular}[b]{c | c | c | c}
%\hline $\epsilon \backslash \, \alpha$ & 0.025 & 0.05 & 0.1  \tabularnewline
%\hline
%0.05 & $8.148 \pm 0.089$ & $7.94 \pm 0.28$ & $7.93 \pm 0.12$  \tabularnewline
%\hline
%0.1 & $8.00 \pm 0.13$ & $\textbf{7.77} \pm 0.23$ & $\textbf{7.77} \pm 0.17$  \tabularnewline
%\hline
%\end{tabular}
%\caption{Top graphs: (left and middle) Loss as a function of the number of model evaluations for the RL scheme with different choices of parameters, (right) the specific actions (samplers) selected by the RL scheme with parameters $\epsilon=0.1$ and $\alpha=0.1$ during the 900 epochs of a calibration for each of the 3 independent runs, to be read from left to right, from top to bottom, note that each epoch provides 4 model evaluations.
%%
%Bottom table: Means and standard errors of the lowest losses achieved by the RL scheme. 
%%
%These results can be compared directly with those of Figure~\ref{fig:convergence} as they have identical ranges on both $x$ and the $y$ axes. }
%\label{fig:convergence-rl}
%\end{figure*}

% In \emph{offline-learning}, the agent learns from previously executed calibrations of se, and \emph{online-learning}, where the agent actively influences the calibration process, and learns from the received feedback.

% \subsection{Offline experiments}
%
\vspace{0.2cm}
\noindent
\textbf{Offline experiments.}
In this section, we train a MAB agent over past calibration histories. 
More precisely, we take the single methods and couples of methods calibrations of Section~\ref{sec:benchmarks}, and process them as if they were observed by a MAB algorithm. 
This approach gives us an estimate of the expected gain of each sampler, and therefore information about the effectiveness of the sampler methods on the specific calibration task.

In the context of MAB solutions, \emph{action-value methods} are methods for estimating the values of actions and for using the estimates to make action selection decisions~\cite {sutton2018reinforcement}. 
%
%Let $Q_t$ be the action-value function at time step $t$, and $Q_t(a)$ the value of action $a$ at time step $t$.
Let $Q(a)$ be the value of action $a$ or, in our context, the value of using a specific search method during a calibration.
One natural way to estimate such values is by averaging the rewards actually received
\begin{equation}
Q(a) = \frac{\sum\limits_{t=1}^{N} R_t \cdot \mathds{1}_{A_t=a}}{\sum\limits_{t=1}^{N} \mathds{1}_{A_t=a}},
\label{eq:q-value-estimate}
\end{equation}
%
% \begin{equation}
% Q_t(a) \doteq \frac{\text{sum of rewards when action } a \text{ taken prior to } t}{\text{number of times action } a \text{ taken prior to } t} = \frac{\sum\limits_{i=1}^{t-1} R_i \cdot \mathds{1}_{A_i=a}}{\sum\limits_{i=1}^{t-1} \mathds{1}_{A_i=a}}
% \end{equation}
%
where $A_t$ is the action chosen at step $t$.
This approach is often called the \emph{sample-average} method~\cite{sutton2018reinforcement}.

\begin{figure*}[t]
\centering
\begin{minipage}{.64\textwidth}
\includegraphics[width=1.0\columnwidth,trim={0.35cm 0 0.35cm 0},clip]{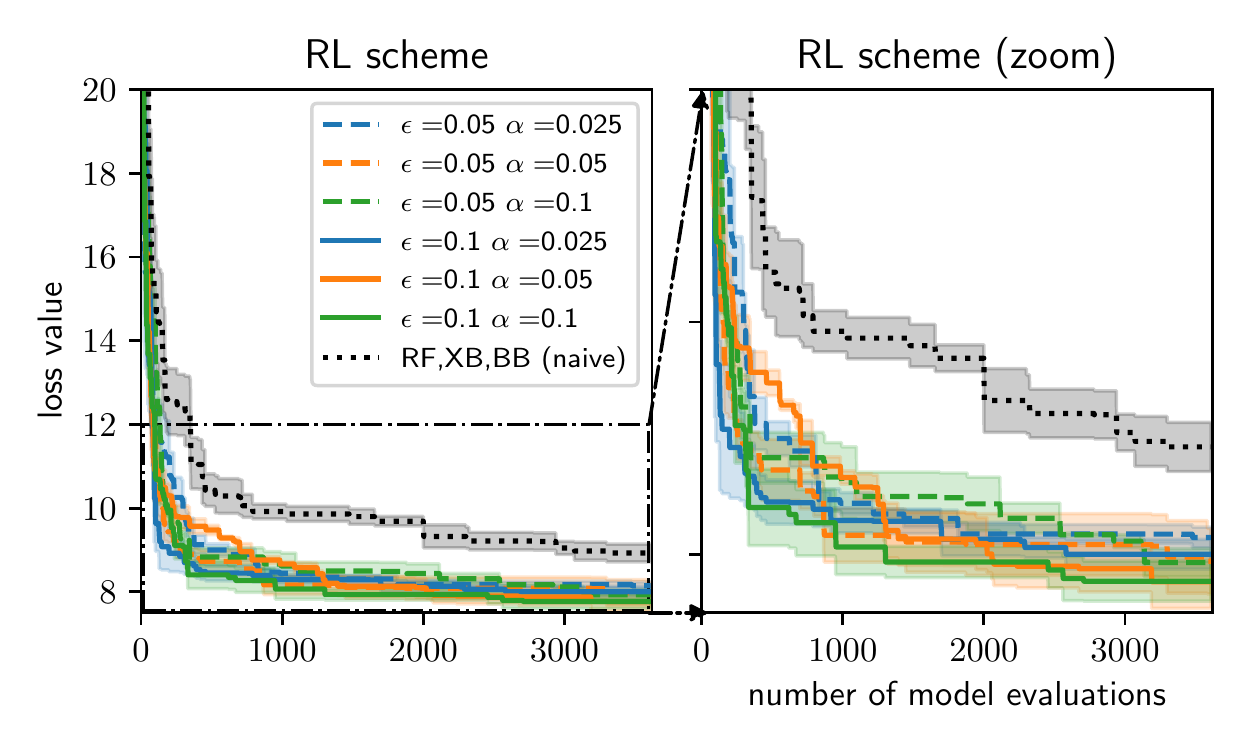}

\centering
\small
\begin{tabular}[b]{c | c | c | c}
\hline $\epsilon \backslash \, \alpha$ & 0.025 & 0.05 & 0.1  \tabularnewline
\hline
0.05 & $8.148 \pm 0.089$ & $7.94 \pm 0.28$ & $7.93 \pm 0.12$  \tabularnewline
\hline
0.1 & $8.00 \pm 0.13$ & $\textbf{7.77} \pm 0.23$ & $\textbf{7.77} \pm 0.17$  \tabularnewline
\hline
\end{tabular}
\end{minipage}
\hfill
\begin{minipage}{.32\textwidth}
\centering
\includegraphics[width=1.0
\linewidth,trim={0 0 0 0},clip]{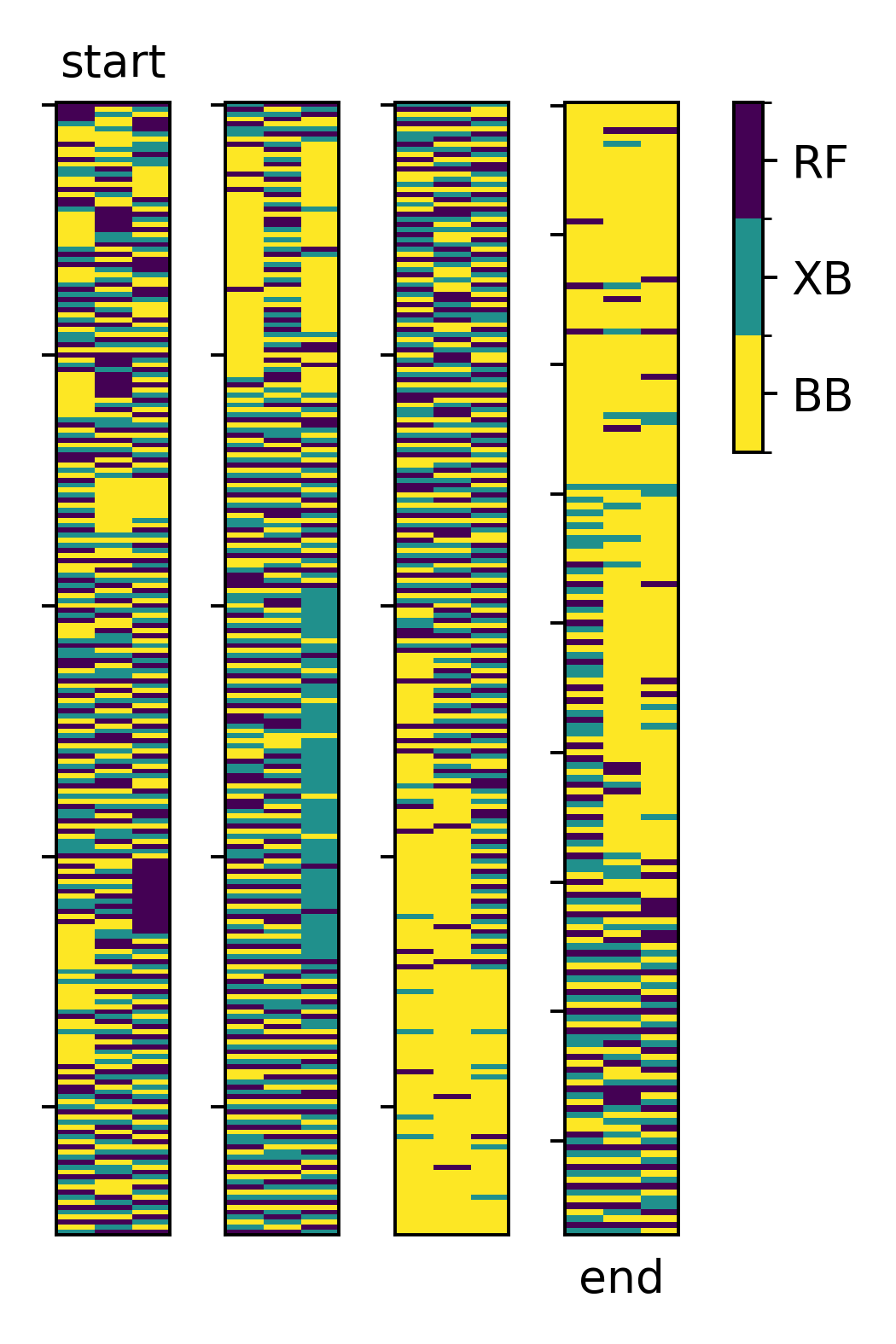}
\end{minipage}
\caption{Top graphs: (left and middle) Loss as a function of the number of model evaluations for the RL scheme with different choices of parameters and for a naive alternation of the samplers, (right) the specific samplers (`actions') selected by the RL scheme with parameters $\epsilon=0.1$ and $\alpha=0.1$ during the 900 epochs of a calibration for each of the 3 independent runs, to be read from left to right, from top to bottom, note that each epoch provides 4 model evaluations.
Bottom table: Means and standard errors of the lowest losses achieved by the RL scheme. 
These results can be compared directly with those of Figure~\ref{fig:convergence} as they have identical ranges on both $x$ and the $y$ axes. }
\label{fig:convergence-rl}
\vspace{-0.2cm}
\end{figure*}

The first two columns of Table~\ref{tab:rl-experiment-offline} provide the results of this analysis when only the single sampler calibrations are considered (``sing. samp.'' column) and when all calibrations are considered (``glob.'' column).
Not surprisingly, the $\rf$ sampler reaches the highest $Q$ value using both datasets, and the results of the ``sing. samp.'' column replicate the hierarchy of samplers of the first panel of Figure~\ref{fig:convergence}.
Interestingly, the value of the $\bb$ sampler dramatically increases when the combined dataset is used, confirming the analysis carried forward in the last section on the effectiveness of using the $\bb$ sampler in combination with a machine learning surrogate sampler.

The third and fourth columns of Table \ref{tab:rl-experiment-offline} offer additional insight.
In these columns, we restrict the value function estimation of Eq.~(\ref{eq:q-value-estimate}) to actions performed in one of two different `states', characterised by the best loss $L_{\best,t}$ being either above the median (``high $L_{\best,t}$'' column) or below the median (``low $L_{\best,t}$'' column).
Models of this kind, where the actions of a MAB agent depend on one or more states (in this case high/low loss value) are known as \emph{contextual} MABs~\cite{langford2007epoch,li_contextual-bandit_2010}.

The results clearly indicate that when the loss is high (typically at the beginning of the calibration) the optimal action is the $\rf$ sampler, but when the loss is low (typically at the end of the calibration) the optimal action becomes, by far, the $\bb$ sampler.
The $\bb$ sampler proposes small perturbations around low-loss parameter combinations, and hence it can be expected to be particularly effective when the calibration has already reached a good minimum, which can be further explored with this method.

This analysis suggests the design of a mixed search scheme that exploits a machine learning surrogate sampler (say $\rf$ or $\xb$) when the loss is sufficiently high, before switching to the $\bb$ sampler towards the end of the calibration.
However, this specific strategy would not be generally applicable as, on a new calibration task, one would not know in advance the loss values that can be achieved, and hence could not set any loss threshold on the choice of sampler.
In the following section, we show how a MAB agent trained on-the-fly can solve this problem by learning this behaviour, without any prior information, during the course of a single calibration run.

\vspace{0.2cm}
\noindent
\textbf{Online experiments.}
% \label{sec:online-experiments}
In online learning schemes, the agent interacts with the environment through a specific policy $\pi$ while simultaneously optimising the policy.
We propose the use of one of the most well-known algorithms for online learning of MAB agents in non-stationary environments: the $\epsilon$-greedy policy with fixed learning rate~\cite{sutton2018reinforcement}.
In this framework, at each step $t$, with large probability $1-\epsilon$ the agent performs a `greedy' action i.e., it selects the action $a$ with the highest value $Q(a)$, and with small probability $\epsilon$ it selects a purely random action.
We can hence write down the $\epsilon$-greedy MAB policy as follows
\begin{equation}
    \pi_t = \begin{cases} 
                   \argmax_{a} \, Q_t(a) & \text{with probability } 1-\epsilon \\
                   \text{random action} & \text{with probability } \epsilon
                  \end{cases}.
\label{eq:eps-greedy}
\end{equation}
After the selected action $a$ is performed, the agent receives a reward $R_t$, and updates the value $Q_t(a)$ as
\begin{equation}
Q_{t+1}(a) = \alpha R_t + (1 - \alpha) Q_{t}(a),
\label{eq:q-function-fixed-alpha}
\end{equation}
where $\alpha$ is referred to as the \emph{learning rate}.
Note that the above update rule can be seen as an exponentially weighted moving average of the rewards obtained through action $a$.
The exponential weighting guarantees that the current value of the $Q$ function is not substantially affected by rewards received many steps earlier and, in turn, this allows the algorithm to adapt to changes of the environment on-the-fly during a calibration.

Figure \ref{fig:convergence-rl} shows the results obtained when using the described scheme with a set of possible actions given by the tree samplers $\rf$, $\xb$ and $\bb$.
The left and middle panels of the figure can be directly compared with the graphs in Figure \ref{fig:convergence}, as they have identical ranges on both $x$ and $y$ axes.
We see that the RL scheme proposed strongly outperforms any other method, or method combination, tested in the previous section.
This happens for all values considered for the parameters $\epsilon$ and $\alpha$, with the best results --by a very narrow margin-- obtained with $ \epsilon = \alpha = 0.1 $.
For comparison, the figure also reports --in a black dotted curve-- the loss achieved by combining the three samplers $\rf$, $\xb$ and $\bb$ in a simple (`naive') alternation.
Such a simple method alternation, with no use of RL, can be imagined to provide a lower-bound on the performance of the RL scheme, and it is seen to give rise to a significantly slower convergence.

The right panel of Figure~\ref{fig:convergence-rl} helps us build intuition around the performance of the RL scheme proposed.
It depicts with different colours the different actions (samplers) selected during the 3 RL calibration runs performed with parameters $ \epsilon = \alpha = 0.1 $.
At the beginning of the calibration (say, the first two columns), the agent explores the different strategies by alternating between the 3 samplers and sometimes exploits a specific sampler with long streaks of identical sampler choices.
Towards the end of the calibration (say, the last two columns), when the loss is low, the agent instead more decisively exploits the $\bb$ sampler, in agreement with the offline experiments described earlier and summarised in Table~\ref{tab:rl-experiment-offline}.

In conclusion, we find that modelling the calibration process as an online learning MAB problem, with actions being given by different available search methods, allows to detect the most promising search methods during the course of a single calibration.
This gives rise to a very efficient scheme, and represents a practical tool to intelligently combine different search methods.

We also explored more sophisticated MAB learning schemes including `kl-UCB'~\cite{garivier2011kl}, `Exp3'~\cite{bubeck2012regret}, and `Discounted Thompson' sampling~\cite{raj2017taming}, but found no substantial improvements. 
Hence, we believe that the $\epsilon$-greedy --fixed learning rate-- algorithm described should be preferred both because of its simplicity and because of its  great computational efficiency which, in turn, implies the absence of any overhead in using the RL scheme over naive method combinations.

\begin{figure*}[!ht]
\advance\leftskip-0cm \includegraphics[width=.8
\linewidth]{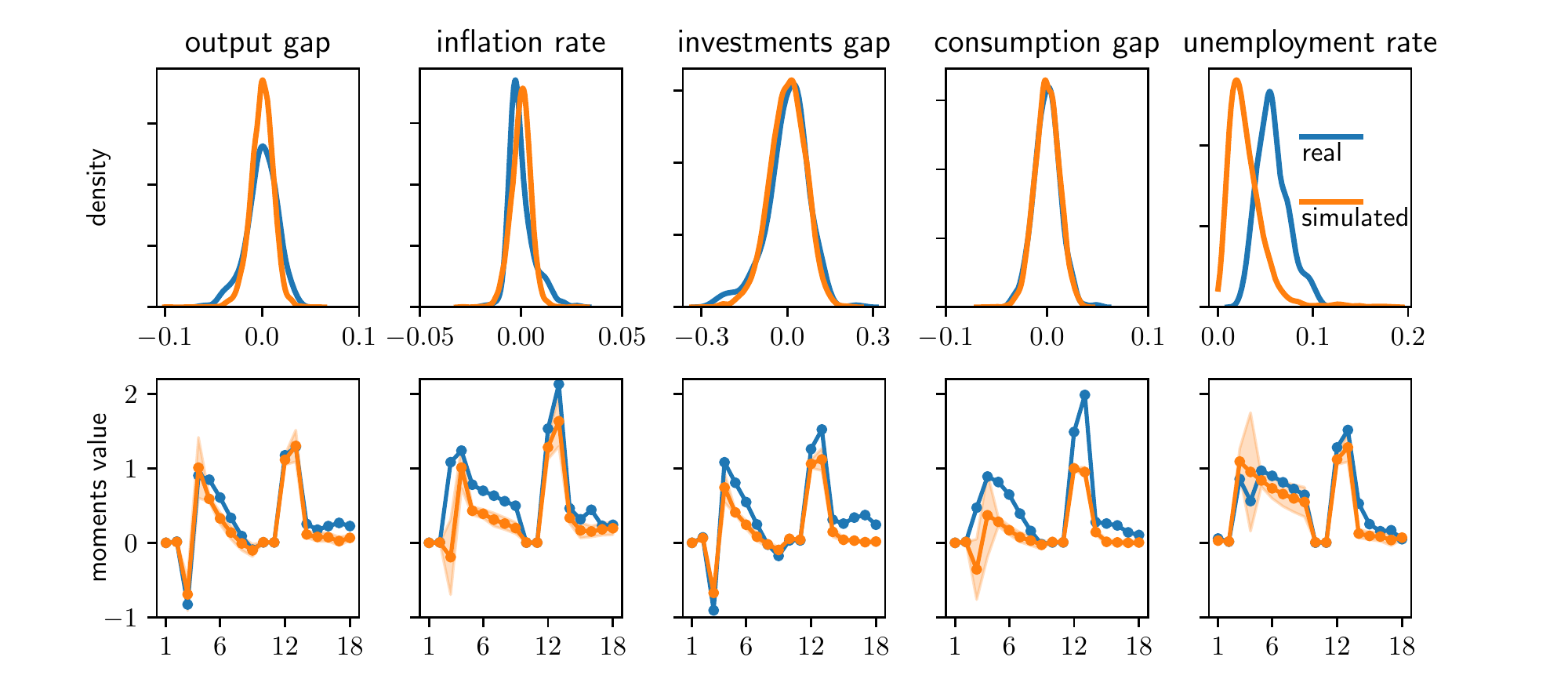}
     \vspace{-15pt} 
    \caption{
    A comparison between distributions and moments of the real series (blue) and the simulated series of lowest loss (orange).
    The first row reports density estimates obtained via a kernel density estimator.
    The second row reports the value of the moments.
    In the second row, the indices from 1 to 18 on the $x$-axis represent the following statistics. 1-4): mean, variance, skewness and kurtosis, 5-9): autocorrelations of increasing time lags. 10-14): mean, variance, skewness and kurtosis of the differentiated time series, 14-18): autocorrelations of the differentiated time series.
}
    \label{fig:density-moments}
    \vspace{-0.2cm}
\end{figure*}

\section{Validation}
\label{sec:validation}

\noindent
\textit{Calibrated model.}
We here verify that the calibrated model is able to approximately reproduce the behaviour of the five variables tracked in the real dataset.
This can be immediately seen by analysing Figure \ref{fig:density-moments}, in which the distribution and the moments of the simulated series with the lowest loss are compared with those computed for the real historical series.
In agreement with \cite{gatti2020rising}, output, consumption and investment are very well captured by the CATS model, while stronger deviations can be observed in inflation and unemployment rates.
Also in agreement with \cite{gatti2020rising} we find that, in general, the CATS model can only partially account for the persistence of the real time series.
This is clear from the fact that the simulated series have systematically lower values of virtually all autocorrelations considered (indices 5-9 and 14-18 in the second-row graphs).
It is important to note here that the calibration experiments performed in this work do not incur into problems of overfitting since the number of parameters and the flexibility of the ABM considered are limited in comparison to the details of the real time series analysed.
This fact is evident from the discrepancy between the high order moments of the real and simulated series in Figure \ref{fig:density-moments}, as well as from the fact that the loss never reaches zero in Figures~\ref{fig:convergence} and~\ref{fig:convergence-rl}.

\begin{figure*}[!ht]
\centering
\advance\leftskip-1cm \includegraphics[width=.6\linewidth,trim={0.35cm 0.2cm 0.35cm 0.35cm},clip]{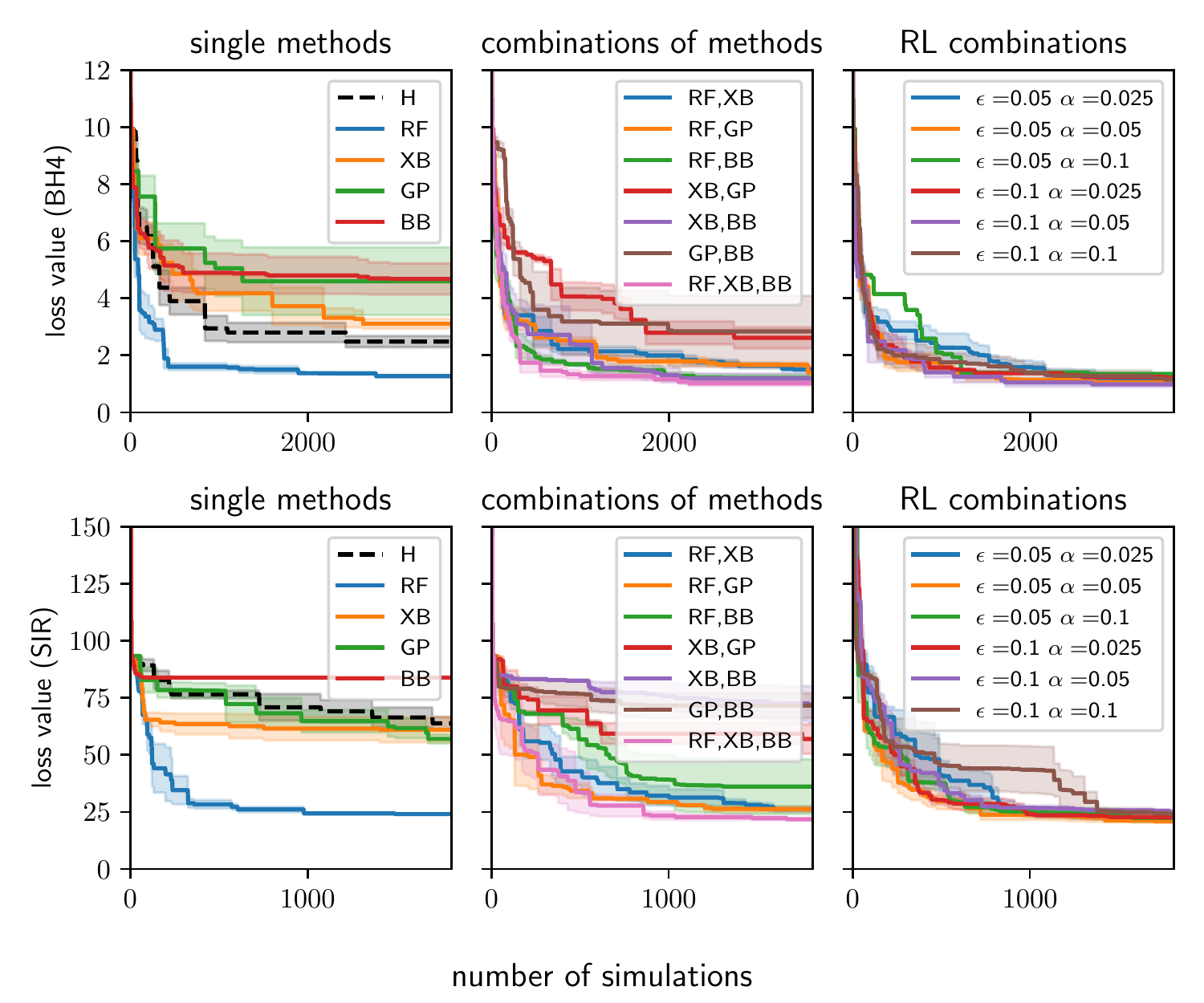}

\centering
\small

\caption{Loss as a function of the number of model evaluations for single methods, simple method combinations, and RL method combinations, for the BH4 model \cite{brock1998heterogeneous} with a method of moments loss (1st row) and for the SIR model with a Euclidean distance loss (2nd row).}
\label{fig:bh4-sir}
\vspace{-0.2cm}
\end{figure*}

\noindent
\textit{Different models.}
%
% This study aims to address the challenge of calibrating a medium-scale macroeconomic ABM.
% %
While the focus of the current work is on the calibration of the CATS model, we verified that our findings hold also for other ABMs.
Specifically, in Figure \ref{fig:bh4-sir} we show the results of similar calibration experiments for two other toy models: the paradigmatic `Brock and Hommes' asset pricing model \cite{brock1998heterogeneous} with a method of moments loss, and a SIR model on a small-world network topology \cite{simoes2008stochastic} with a Euclidean distance loss.
In agreement with the rest of this work, we find that the $\rf$ sampler is the best-performing sampler when methods are used in isolation, that coupling different samplers generally provides better performances, and that our RL-scheme can be successfully used to intelligently combine search methods.
However, these alternative calibration tasks are much simpler than the calibration of the CATS model considered in the rest of this work.
For this reason, we do not see a significant performance improvement in using RL combinations over simple combinations but, importantly, we consistently find the performance of the RL-scheme to be as good as the best samplers or sampler-combinations tested, without requiring any trial and error.

\noindent
\textit{Reproducibility.}
In the interest of reproducibility an easy-to-use implementation of the reinforcement learning scheduler proposed in this work has been released in open source
within the \emph{Black-it} package\footnote{A Jupyter notebook to experiment with it is available at \url{https://github.com/bancaditalia/black-it/blob/main/examples/RL_to_combine_search_methods.ipynb}}.

\section{Conclusions}
\label{sec:conclusion}

In this work, we systematically compare the performance of 5 search strategies, taken in isolation and in combination, on a method-of-moments calibration of a standard macroeconomic ABM.
Our results show that calibration based on machine learning surrogate samplers, of the kind proposed in \cite{lamperti2018agent} but using a random forest algorithm for interpolation, provides superior performance with respect to the other search methods.
Our results further show that coupling different search methods together gives rise to search strategies that typically improve over their constituents.
The empirical efficacy of random forest search methods and of combining different search methods can be of practical help to researchers interested in calibrating and using medium and large-scale ABMs.
However, when combining different search methods a natural issue arises about which methods should be combined, and in which way. 

We provide a solution to this issue by framing the choice of search methods as a multi-armed bandit problem, and leveraging a well-known reinforcement learning scheme to select the best method on-the-fly during the course of a single calibration.
The RL scheme proposed outperforms any other method or method combination tested, and thus provides a practical tool for researchers interested in efficiently calibrating ABMs.

In the future, it would be interesting to deepen the analyses of the present study in two possible lines of research, based on either extensions of the banchmarking experiments of Section~\ref{sec:benchmarks} or on further investigations into the RL scheme of Section~\ref{sec:reinforcement-learning}.

The benchmarking framework could be extended in several dimensions.
The first is the testing of other standard search methods, such as particle swarm samplers or machine learning samplers based on neural networks. 
The second is the inclusion in the analysis of other measures of goodness of fit, in addition to the method of moments, such as likelihood measures, Bayesian measures \cite{grazzini2017bayesian,dyer2022black}, or information theoretic measures \cite{lamperti2018information}.
The third is the addition of other widely known macroeconomic ABMs~\cite{dawid2018agent} to the analysis, such as the so-called ``K+S'' model \cite{dosi2010schumpeter}, or the recent large-scale model of \cite{poledna2023economic}. 
This would allow quantitative benchmarking not only of the calibration strategies, but also of the different models when calibrated on the same data.
The final direction would involve appropriately increasing the data on which the ABMs are calibrated and tested, potentially with more variables and with more national economies.
In essence, while the present work is an important step towards a systematic assessment calibration methods for medium and large-scale economic ABMs, all of the above mentioned directions would surely represent equally important steps towards an increasingly more data-driven ABM development.

The RL scheme proposed may also deserve further specific investigation. 
For example, one could verify whether the RL search method developed here maintains its high performance also in the more general setting of black-box function optimisation, perhaps in other specific application domains that might have peculiarities similar to the ABM calibration problem.
One might also try to extend the simple (yet effective) MAB framework introduced here, by providing more `contextual' information to the agent and hence attempting to represent the ABM calibration problem either as an online contextual-MAB problem, or directly as a partially-observable MDP \cite{kaelbling1998planning}.
%
% Potentially, the problem could even be made suited for a pure MDP formulation by feeding the entire history of the past sampled point to the agent that needs to decide on the next search method, or directly decide the specific points to sample as proposed in \cite{chen2017learning}.

%Acknowledgements
%\subsection*{Acknowledgements}
%\acknowledgements
\begin{acks}
D.C. acknowledges funding from the European Union’s Horizon 2020 research and innovation programme under the Marie Skłodowska-Curie grant agreement No 956107, ``Economic Policy in Complex Environments'' (EPOC).

We would like to thank Marco Pangallo (CENTAI institute), Herbert Dawid (Bielefeld University), Bence Mér\H{o} (Bank of Hungary), Paolo Pellizzari (University of Venice) and the anonymous reviewers of the ICLR workshop ‘AI4ABM’, of the AAAI bridge program ‘AI for Financial Institutions’ and of ICAIF'23, for constructive feedback on this work.

The views and opinions expressed in this paper are those of the authors and do not necessarily reflect the official policy or position of Banca d’Italia.

\end{acks}
% ?? Anonymous reviewers of ECAI ??

% We would like to thank the referees for their comments, which helped improve this paper considerably.

\balance

% Bibliography
\bibliographystyle{ACM-Reference-Format.bst}

\bibliography{acm-ec, RL-refs}

\end{document}